\documentclass[twocolumn]{article}
\usepackage{graphicx}
\usepackage{booktabs}
\usepackage[backend=bibtex,style=ieee]{biblatex}
\usepackage{hyperref}
\title{\bf Real-time Neural Rendering of LiDAR Point Clouds}

\author{
    Joni Vanherck$^{*1}$, 
    Brent Zoomers$^{*1,2}$, 
    Tom Mertens$^{1}$,
    Lode Jorissen$^{1}$, 
    Nick Michiels$^{1}$
}

\date{
    {\small
        $^1$Hasselt University, Digital Future Lab \& Flanders Make \\
        $^2$Research Foundation - Flanders \\
        $^*$Equal contribution
    }
}

\begin{document}

\maketitle

\begin{abstract}
Static LiDAR scanners produce accurate, dense, colored point clouds, but often contain obtrusive artifacts which makes them ill-suited for direct display. We propose an efficient method to render photorealistic images of such scans without any expensive preprocessing or training of a scene-specific model. A naive projection of the point cloud to the output view using 1$\times$1 pixels is fast and retains the available detail, but also results in unintelligible renderings as background points leak in between the foreground pixels. The key insight is that these projections can be transformed into a realistic result using a deep convolutional model in the form of a U-Net, and a depth-based heuristic that prefilters the data. The U-Net also handles LiDAR-specific problems such as missing parts due to occlusion, color inconsistencies and varying point densities. We also describe a method to generate synthetic training data to deal with imperfectly-aligned ground truth images. Our method achieves real-time rendering rates using an off-the-shelf GPU and outperforms the state-of-the-art in both speed and quality.
 
\end{abstract}  
\section{Introduction}

Despite the tremendous progress in novel view synthesis from RGB photos, LiDAR offers some compelling advantages: accuracy, high density, ability to operate in dark conditions, large area coverage, to name a few. Modern scanners also output colored points, which are estimated using a spherical camera array embedded in the scanner. Such colored clouds can be rendered directly using standard graphics rendering. But such renderings are also prone to specific artifacts. Scans are typically composed of multiple point clouds captured originating from different scanner positions within a scene, leading to the following problems:
\begin{itemize}
   
    \item \textbf{Incomplete coverage}: even with multiple viewpoints, LiDAR scans contain significant gaps when foreground objects occlude the background, and because the scanner does not fully cover the vertical field-of-view.
    \item \textbf{Color inconsistencies}: When point clouds from different scanner positions are merged, view-dependent lighting effects such as reflections and, imaging and post-processing defects can cause color discrepancies between different views. 
    \item \textbf{Density variations}: The density of points decreases with distance from the scanner, leading to a wide variety of sampling densities in the merged point clouds.
\end{itemize}

One could imagine preprocessing the point cloud to obtain a higher quality version without said artifacts and rendering it instead. But this would imply a tremendous input-sensitive cost before the cloud can be displayed, given the enormous size which can easily approach dozens or hundreds of millions of points. Instead, we are interested in a real-time method that is mainly output-sensitive. To this end, we project the points directly to the output view as 1$\times$1 pixels. The only required preprocessing to mitigate severe input-sensistivity is a fast spatial partitioning using a simple uniform grid to enable view frustum culling.

\begin{figure*}[htb]
    \centering
    \includegraphics[width=\linewidth]{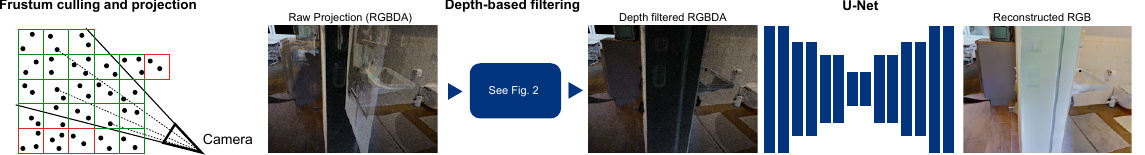}
    \caption{This figure provides an overview of the proposed method. A space-partitioned point cloud is projected onto the camera view using frustum culling, generating RGB, depth (D) and binary fill mask (A). The depth channel is used for depth filtering, as detailed in Fig. \ref{fig:depthfiltering_step}. A fully convolutional U-Net then processes the resulting depth-filtered RGBDA image to produce the final output.}
    \label{fig:overview}
\end{figure*}

Unfortunately, a naive point projection produces unintelligible renderings as background points leak in between the foreground pixels (see leftmost image in Fig. \ref{fig:overview}). We address this via a real-time neural image reconstruction method based on a classic convolutional U-net. This not only produces photorealistic results, but also mitigates the aforementioned LiDAR artifacts. We found that filtering out leaked background points before the image is passed to the U-net increases perceptual quality and detail. These background points do not bear any relevance w.r.t. reconstructing the foreground, and reduce the task of the reconstruction U-net to inpainting the gaps aside from mitigating LiDAR artifacts. We implement this filter using a depth-guided heuristic. We also describe a method to generate synthetic training data to deal with imperfectly-aligned ground truth images.


\begin{figure*}[htb]
    \centering
    \includegraphics[width=\linewidth]{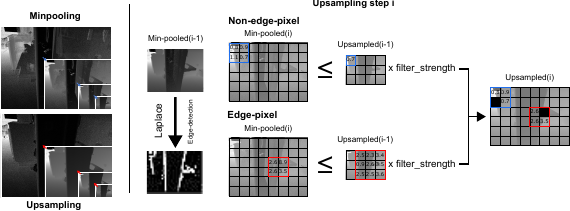}
    \caption{On the left, the hierarchical structure is illustrated, where we use minpooling with kernel size 2$\times$2 and stride 2 to downsample the image n times. Upsampling then occurs until the original image size is reached. On the right, an upsampling step is depicted. First, a Laplacian kernel is applied to the min-pooled(i-1) image for edge detection. If the pixel being upsampled in upsampled(i-1) is a non-edge pixel, its value is compared to the corresponding four pixels in min-pooled(i). For edge pixels, these four values are also compared with the eight neighboring pixels in upsampled(i-1). A filtering factor, filter\_strength, determines how many points are retained, with smaller values producing sparser results. This process generates an upsampled image, where unfiltered pixels remain empty. If this is not the final upsampling step, empty pixels are filled using linear interpolation from upsampled(i-1).}
    \label{fig:depthfiltering_step}
\end{figure*}

\section{Related Work}
Point cloud rendering is a decades-old research topic, that departs from the classic notion of connected primitives such as triangles, and instead displays surfaces directly as points. Due to space limitations, we will only discuss recent work that focuses on leveraging deep learning.

\textbf{Neural Point Cloud Rendering} The majority of methods train neural feature vectors on a per scene basis~\cite{hu2023trivolpointcloudrendering}, or require encoding them using a pretrained network~\cite{Rakhimov_2022_CVPR}. Rendering involves projecting the feature-enriched points and reconstructing them using CNNs, typically a U-net. While some methods obtain features from the point cloud itself~\cite{hu2023point2pixphotorealisticpointcloud}, others rely on a set of registered photographs from the same scene~\cite{npbg,franke2024trips,Rakhimov_2022_CVPR}. Because each point requires such features, these methods do not scale well with input size, which can easily run into dozens of millions of points for LiDAR scans. Our method foregoes any form of scene-specific training or encoding by directly rasterizing colored points into the output view, requiring only a U-net to reconstruct the final image.

\textbf{Filtering} Our depth-guided filter is designed to discard unwanted background information. A similar problem has been addressed in InvSfM~\cite{pittaluga2019revealing}, where the authors train a dedicated U-net to resolve foreground visibility, but this comes at the cost of increased inference time and requires a suitable ground truth dataset. NPBG++~\cite{Rakhimov_2022_CVPR} employs a depth heuristic similar to ours to transfer input image features to aggregated features on the point cloud. Theirs operates on only one level that has to set based on the chosen dataset, whereas ours is multi-resolution.

\textbf{Direct Rendering} Pointersect~\cite{pointersect} renders colored point clouds based on a model that is trained to predict ray intersection tests. This approach is scene agnostic and works on different types of point clouds, such as RGBD and LiDAR. The main drawback of this technique is its inference time, as rendering one image can take up to several minutes. It also does not address the difficulty of mitigating defects and gaps seen in LiDAR scans.

\section{Method}
The different steps of our method are: projecting the points to the output view, filtering the points using a depth-based heuristic, final reconstruction with a U-net and generating training data for the latter. The following paragraphs detail these steps. See Fig. \ref{fig:overview} for an overview.

\textbf{Point Cloud Projection} We first project the points from the point cloud to the desired view. We implement a simple frustum culling approach based on a lightweight space partitioning method in the form of a uniform grid to achieve constant time complexity at render time. This only takes 30ms per 1M points, measured with our CPU implementation on a high-end desktop. After culling, each point is rasterized as a 1$\times$1 to preserve the maximum amount of detail. Soft z-buffering resolves collisions by averaging the colors of points that arrive at similar depths, given a small depth threshold. In regions where pixels on the front do not cover the full area, the background may leak into the foreground. While using larger splats can reduce this leakage, it also obscures detail in dense regions. To resolve this issue, we propose a heuristic depth filter in the next paragraph.

\textbf{Depth filtering} We propose a heuristic based on hierarchical min-pooling. Each min-pooling step removes background points. We then upsample each min-pooled level starting from the coarsest level back to the original resolution. Each step restores the original foreground pixels while discarding the ones from the background. The key idea is that for any group of 2$\times$2 pixels at a given level, we can discern between foreground and background using a threshold based on the corresponding 1$\times$1 depth at the parent level. We refer to Fig. \ref{fig:depthfiltering_step} for more info, and the supplementary material for implementation details.
 
\textbf{U-Net Reconstruction}
The original U-Net~\cite{unet} focused on segmentation, where the goal was to segment biomedical images. Other works have shown that this architecture performs well for other image-processing tasks, such as denoising~\cite{gurrola2021residual}. We modify the U-Net to take RGBDA as input, where D represents z-buffer depth, and A (alpha) indicates whether a point was projected on this pixel. Points that were discarded by our depth-guided filtering will also have a zero alpha value. Instead of using cross-entropy loss combined with pixel-wise soft-max, we take inspiration from novel view synthesis~\cite{kerbl3Dgaussians} and combine $L_1$loss and a perceptual similarity loss to balance pixel-level accuracy with perceptual similarity. 
\[L_{final} = \lambda_{1}*L_1 + \lambda_{2}*LPIPS,\]
where $\lambda_1 = 0.9$ and $\lambda_2 = 0.1$. We also added Batch normalization after each convolutional step during downsampling to accelerate training. We trained our network for 170 epochs on 36 ScanNet++ v1 scenes~\cite{yeshwanthliu2023scannetpp}, consisting of 19,188 training image pairs and 4,798 test image pairs, where each pair consists of a ground-truth image and a rendered LiDAR point cloud image. We used random crops of $320 \times 320$ pixels to reduce memory usage during training.

\begin{figure*}
    \centering
    \includegraphics[width=\textwidth]{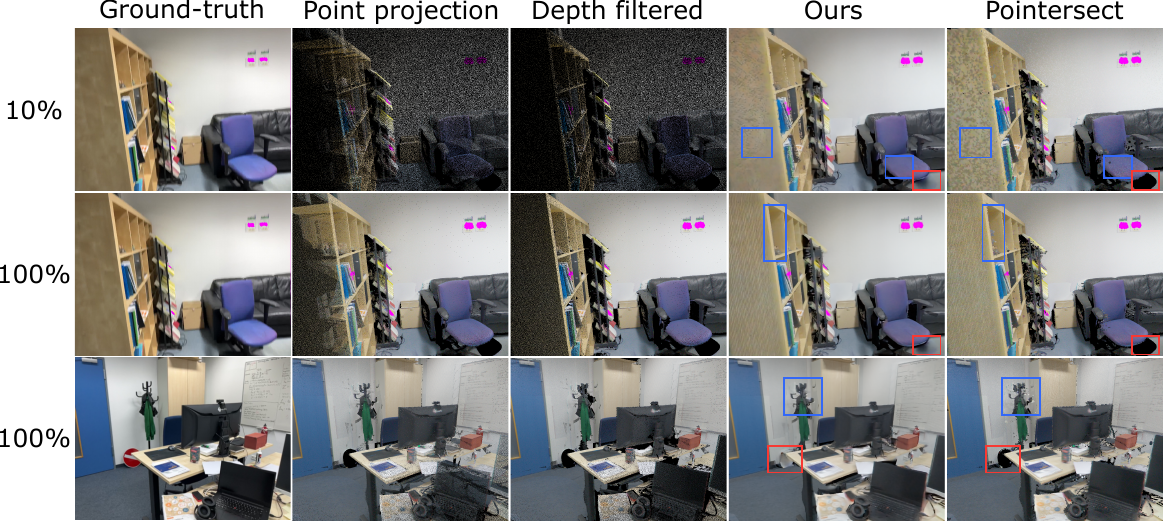}
    \caption{This figure illustrates the steps in our method and compares with Pointersect~\cite{pointersect}. We show a low-density scenario by decimating the point cloud to 10\% and a high-density scenario. In both cases, our method produces a more naturally looking image overall w.r.t. Pointersect. Specifically, Pointersect is not able to fill large and small gaps (indicated in red), and exhibits more noise (indicated in blue).}
    \label{fig:results}
\end{figure*}

\textbf{Training Data Generation}
Our network was initially trained on the ScanNet++ v1 dataset, which provides LiDAR scans with registered camera poses. However, we found that these poses do not produce pixel-perfect registration, making it challenging to teach the network to produce sharp details. We therefore generate a synthetic dataset based on the ScanNet++ v1 data by modifying ground truth registered photos to make them visually similar to a raw point cloud projection. The most important aspect is the effect of missing points due to point cloud sparsity and discarded points after depth-guided filtering. To this end, we project the point cloud from the slightly misaligned pose corresponding to the ground truth image, but only retain the depth $D^s$ and alpha channel $A^s$. The ground truth's $(RGB)^{gt}$ pixels of the ground truth are zeroed where $A^s$ is zero, yielding $(RGB)^s$, such that the final training pair becomes $[(RGBDA)^s, (RGB)^{gt}]$. To further improve generalization, we simulate multiple scanner poses by applying random brightness and contrast variations. See supplementary material for details.

We also trained a version of our U-net that does not expect a depth-based filtered input. To generate the synthetic training data we modify the above approach to include background leakage from the point cloud in the training images. This requires deciding whether a filled pixel ($A^s=1$) is part of the foreground or background, which is made by our depth-guided filtering heuristic. Foreground pixels are then set to $(RGB)^{gt}$, while background ones receive $(RGB)^{s}$. Pose misalignment on the latter pixels is less of a concern as the network should learn to ignore them.

\begin{table}[h]
\caption{This table compares our method to Pointersect~\cite{pointersect} at two resolutions: 1920$\times$1440 and 960$\times$720, using full-size point clouds. We evaluate both methods with PSNR, LPIPS, SSIM, and rendering speed. Our method outperforms Pointersect in image quality and is over 2000 times faster in rendering.}
\label{tab:speed_points_summary}
\centering
\resizebox{\columnwidth}{!}{
\begin{tabular}{lcccc}
\toprule
Method & PSNR$\displaystyle \uparrow$ & LPIPS$\displaystyle \downarrow$ & SSIM$\displaystyle \uparrow$ & FPS$\displaystyle \uparrow$ \\
\midrule
Ours {\scriptsize (1920$\times$1440)} &  16.10 & 0.34 & 0.74 & \textbf{13.63} \\
Pointersect {\tiny (1920$\times$1440)} & 14.94 & 0.49 & 0.60 & 0.006 \\
\midrule
Ours {\scriptsize (960$\times$720)} & 16.29 & 0.34 & 0.71 & \textbf{49.71} \\
Pointersect {\tiny (960$\times$720)} & 14.97 & 0.48 & 0.55 & 0.021 \\
\bottomrule
\end{tabular}
}
\end{table}

\section{Experiments}
Our method is implemented in CUDA, allowing for real-time rendering of point clouds. All experiments were run on an NVIDIA GeForce RTX 4090 and 13th Gen Intel(R) Core(TM) i9-13900KF. Training took around 60 hours on the same machine. We evaluated our method on a random subset of 55 scenes from ScanNet++ v2 and removed them from our training set. For comparison, we evaluate our method against Pointersect~\cite{pointersect}. The point clouds have an average size of 23 million points without any decimation. Results are shown in Table \ref{tab:speed_points_summary} and Fig. \ref{fig:results}. We note that metrics in Table \ref{tab:speed_points_summary} are subject to slightly inaccurate poses of ground truth images in ScanNet++, and the color distribution of the ground truth images the point colors of the LiDAR scanner do not align because they originate from different cameras. Still, we present them here due because we believe they provide a reasonable indication of quality. We also compared reconstruction with and without the depth filter and noticed little to no difference in quality metrics. However, we notice a clear difference upon visual inspection. Using the depth filter tends to produce sharper details. It also includes more high-frequency, noisy artifacts than without the depth-guided filter. The latter does better at filling holes but tends to produce a more blurred and washed-out appearance. We find that using the depth filter results in clearer, more apprehensible images, especially for scenes with high depth complexity where background leaking is more prominent. A visual comparison can be found in the supplementary material.

\section{Discussion \& Conclusion}
We have shown that our approach can render images from LiDAR-based point clouds in real-time by combining our depth heuristic with a compact U-Net, and ensuring perfectly registered training data pairs using synthetic rendering. We have compared our approach to the state-of-the-art and show that our approach achieves higher quality images and 3 orders of magnitude speedup.

Our approach addresses a subset of challenges in the realm of neural point cloud rendering of static LiDAR data. Our work focuses on obtaining real-time rendering rates with relatively simple implementation and modest training requirements. Other challenges, such as large-scale defects in point clouds due to reflective, transparent, or moving objects still need to be addressed. Some of these artifacts can be seen in the supplementary video.

\textbf{Acknowledgements}
This research was partly funded by the Special Research Fund (BOF) of Hasselt University (R-14360, R-14436) and the FWO fellowship grant (1SHDZ24N), the European Union (HORIZON MAX-R,
Mixed Augmented and Extended Reality Media Pipeline, 101070072), the Flanders Make’s XRTwin SBO project (R-12528). This work was made possible with support from MAXVR-INFRA, a scalable and flexible infrastructure that facilitates the transition to digital-physical work environments.
\printbibliography 
\end{document}